\documentclass[letterpaper, 10 pt, conference]{ieeeconf}  

\IEEEoverridecommandlockouts                              

\usepackage{amsmath,amsfonts}
\usepackage{amsmath}

\usepackage{algorithmic}
\usepackage{algorithm}
\usepackage{array}
\usepackage{textcomp}

\usepackage{stfloats}
\usepackage{url}
\usepackage{verbatim}
\usepackage{graphicx}
\usepackage{cite}

\usepackage{stfloats}
\usepackage{url}
\usepackage{verbatim}
\usepackage{graphicx}
\usepackage{cite}

\usepackage[table,xcdraw]{xcolor} 

\usepackage{amsmath}
\usepackage{amssymb}
\usepackage{multirow}
\usepackage{booktabs} 
\usepackage{array} 
\usepackage{pifont} 

\usepackage{xcolor}
\usepackage{soul}

\title{\LARGE \bf
Compliant Beaded-String Jamming For Variable Stiffness Anthropomorphic Fingers
}

\author{Maximilian Westermann$^{1}$, Marco Pontin$^{1}$, Leone Costi$^{1}$, Alessandro Albini$^{1}$, and Perla Maiolino$^{1}$
\thanks{This work was supported by  Engineering and Physical Sciences Research Council (EPSRC) Grant EP/V000748/1}
\thanks{$^{1}$Maximilian Westerman, Marco Pontin, Leone Costi, Alessandro Albini, and Perla Maiolino are with the Oxford Robotics Institute, University of Oxford, Oxford, OX1 2JD, United Kingdom;
{\tt\small maximilian.westermann/marco.pontin/leone.costi}
{\tt\small alessandro.albini/}
{\tt\small 
perla.maiolino@eng.ox.ac.uk}}}

\begin{document}

\maketitle
\thispagestyle{empty}
\pagestyle{empty}

\begin{abstract}
Achieving human-like dexterity in robotic grippers remains an open challenge, particularly in ensuring robust manipulation in uncertain environments. Soft robotic hands try to address this by leveraging passive compliance, a characteristic that is crucial to the adaptability of the human hand, to achieve more robust manipulation while reducing reliance on high-resolution sensing and complex control. Further improvements in terms of precision and postural stability in manipulation tasks are achieved through the integration of variable stiffness mechanisms, but these tend to lack residual compliance, be bulky and have slow response times. To address these limitations, this work introduces a Compliant Joint Jamming mechanism for anthropomorphic fingers that exhibits passive residual compliance and adjustable stiffness, while achieving a range of motion in line with that of human interphalangeal joints. The stiffness range provided by the mechanism is controllable from 0.48$\,$Nm/rad to 1.95$\,$Nm/rad (a 4x increase). Repeatability, hysteresis and stiffness were also characterized as a function of the jamming force. To demonstrate the importance of the passive residual compliance afforded by the proposed system, a peg-in-hole task was conducted, which showed a 60\%\ higher success rate for a gripper integrating our joint design when compared to a rigid one.
\end{abstract}

\section{Introduction}

Soft hands have emerged as a promising solution to the challenges of robotic manipulation. By leveraging morphological computation and passive behaviors provided by their inherent compliance \cite{passive, doi:10.1089/soro.2020.0153, 9761831}, these demonstrate increased robustness in manipulation tasks, being able to compensate for uncertainties and external disturbances with minimal onboard sensing and reduced computational effort \cite{897777, 9761831, annurev:/content/journals/10.1146/annurev-control-060117-105003, junge}. However, their lack of rigidity limits their application in tasks demanding precision or when high-force interactions are involved \cite{Rodrigue2024, stiff}.

\begin{figure}[t]
    \centering
    \includegraphics[width=\columnwidth]{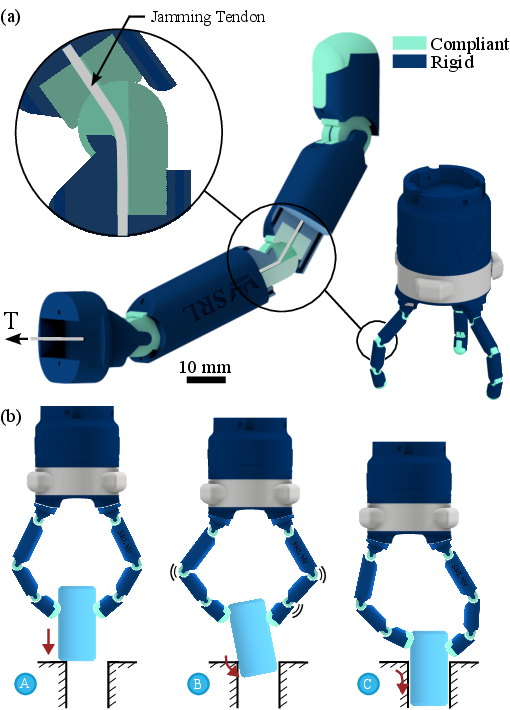}
    \caption{Synoptic view of the proposed Compliant Joint Jamming design. (a) Close-up view of the variable stiffness finger and its internal structure, alongside a three-fingered gripper. (b) Example highlighting the benefits of the passive residual compliance afforded by the Compliant Joint Jamming design. The the fingers passively adapt and allow the peg to orient itself properly without losing their grasp on it, enabling successful completion of the task despite the initial misalignment.}
    \label{fig:first}
    \vspace{-1.5em}
\end{figure}

To mitigate these problems, variable stiffness (VS) mechanisms have been incorporated into soft hands \cite{ Shan2023, Dou}. In \cite{doi:10.1089/soro.2020.0153}, for example, the authors explored the use of shape memory polymers to achieve a variable stiffness soft hand with a lightweight and compact finger shape able to achieve a 5x variation in stiffness. Although promising, systems based on smart materials tend to suffer from slow stiffness adjustment cycles (10$\,$s), preventing their applicability in dynamic tasks requiring rapid adaptive responses \cite{VSpeed, Aydin}. Another popular approach is represented by jamming methods, due to the wide range of stiffness achievable and ease of control. Vacuum-based techniques, in particular, represent a popular choice. In \cite{Gilday}, a granular jamming soft hand was developed imitating the passive dynamic properties of a human hand. In the study the authors also successfully demonstrated the benefits of VS in combination with passive compliance in a pick-and-place task involving spherical objects. However, for these mechanisms, the intrinsic coupling between surface area and achievable stiffness results in bulky joints which, when applied to anthropomorphic fingers, restrict mobility and overall dexterity \cite{tschiersky2024flexure, rotat}. Approaches based on antagonistic actuation have also been explored. In \cite{tend}, for example, the authors exploited a tendon-driven system coupled with a specialized control strategy to mimic the behavior of a human metacarpal joint in a robotic finger. Overall, these approaches can achieve faster and more diverse stiffness adjustments compared to the VS mechanisms analyzed previously, but they often rely on complex and computationally intensive control algorithms which pose significant challenges in implementation \cite{Dou, stiff}.

In contrast, Beaded-String Jamming (BSJ), a tendon-driven system, stands out offering compactness, speed, and simplicity in control \cite{OGBead, bead, Jam, doi:10.1089/soro.2022.0232}. However, it only provides two-states, having either free relative motion of the beads or full rigidity through \textit{bulk}\cite{Jam}, and therefore lacks the ability to control joint stiffness in a continuous range. Most importantly, jamming-based mechanisms available today lack passive residual compliance, meaning the ability of joints to have controllable stiffness while maintaining fully elastic behavior in response to external forces, which we believe to be a key requirement to achieve robust dexterous manipulation.

To address this gap, this study proposes Compliant Joint Jamming (CJJ) as a key development of traditional BSJ that achieves passive residual compliance (Fig. \ref{fig:first}). The proposed mechanism is fast acting, with easily controllable and ranged stiffness (4x increase). Its compactness enables its implementation in anthropomorphic fingers, while its design allows the joints to maintain a range of motion comparable to that of human interphalangeal joints. Having characterized our mechanism for holding torque, stiffness, hysteresis and fatigue, we validated the role of passive residual compliance in achieving more robust manipulation through the success rate in a peg-in-hole task.

In the following, the design and manufacturing of the CJJ fingers is presented in Section \ref{sec:m&m}, followed by the detailed characterization of the behavior of the VS hinge mechanism with respect to the jamming tension applied (Section \ref{sec:res}). The validation experiment and its discussion conclude the empirical section of the paper. Finally, our concluding remarks and future outlook of the research are provided in Section \ref{sec:end}. 

\section{Material and Methods}
\label{sec:m&m}

The design of the CJJ fingers stemmed from the integration of functional aspects, such as the need for compliant and rigid elements, manufacturing constraints and dimensional requirements. Each of these aspects is analyzed in detail in the following subsections.

\subsection{Structural Design}

\begin{figure}[t]
    \centering
    \includegraphics[width=\columnwidth]{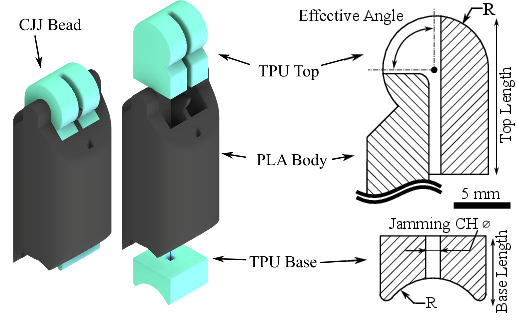}
    \caption{Structure of the variable stiffness finger highlighting the main design parameters. The radius R affects the holding torque of the bead, while the length of the TPU bead elements determines the bending behavior of the joint. The effective angle determines the amount of rotation the phalanges can achieve when not jammed.}
    \label{fig:jjj}
    \vspace{-1em}
    
\end{figure}

\begin{figure}[t!]
    \centering
    \includegraphics[width=\columnwidth]{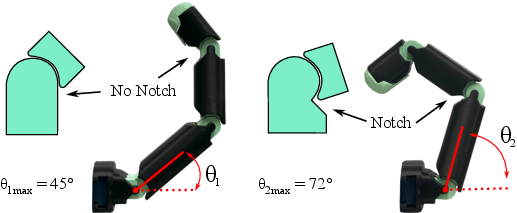}
    \caption{Diagram illustrating how adding a notch to the bead design increases the overall maximum range of motion of the passive finger.}
    \label{fig:not}
    \vspace{-0.5em}
    
\end{figure}

The structure of each phalanx is broken down in Fig. \ref{fig:jjj}. Overall dimensions are based on human anatomy \cite{tschiersky2024flexure}, with each bead measuring 20$\,$mm in length, 10$\,$mm in height and 10$\,$mm in width, resulting in a radius R of 5\,mm. From a kinematic perspective, each joint in the finger provides one degree of freedom, mimicking human anatomy and differentiating itself from the ball and socket joint traditionally used in BSJ. Additionally, a notch cut into the top of the phalanx allows for an extended range of motion of each joint, which reaches 72$^\circ$ (Fig. \ref{fig:not}) bringing it closer to that of a human finger \cite{frange}. 

Material-wise, each phalanx integrates a rigid body together with two compliant elements at the tip and at the base, meant to mesh with corresponding ones on adjacent phalanxes. The 1\,mm steel jamming tendon, which also holds the finger together, is wired through a channel running through the center of each phalanx assembly and a ferrule is used to block it at the tip of the finger, allowing the tension of the tendon to be converted to compression at the joints. Consideration must be put into the selection of the tendon as its material and dimensions can directly affect the base stiffness of the joint and its minimum bending radius, as well as the maximum applicable force. 

\begin{figure}[t]
    \centering
    \includegraphics[width=\columnwidth]{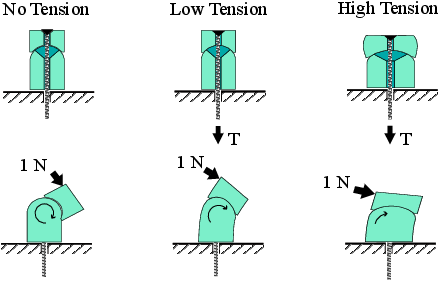}
    \caption{Diagram showing the working principle of the proposed variable stiffness mechanism. When no tension is applied, the two halves of the bead are able to freely rotate relative to each other. The tension applied to the tendon translates into compression between the mating surfaces of the two halves, activating the \textit{bulk} state: the static friction causes the bead to act in unit as a beam undergoing bending. The jamming tension also has the effect of altering the bead geometry, causing the it to contract axially and expand sideways, leading to an increase in its bending stiffness.}
    \label{fig:mech}
    \vspace{-1.5em}
\end{figure}

Fig. \ref{fig:mech} displays the working principle of the proposed mechanism. The combination of jamming tension and deformation of the compliant elements allows for the controllable stiffness range of the joint. With no tension being applied, free relative movement is achieved within the knuckles of the finger. An increase in tension provides friction between the two meshing surfaces of each interphalangeal joint. Within the static friction limit, the joint no longer behaves as a hinge, but rather can be thought of as an elastically deforming beam. An increase in jamming tension produces an increase in static friction and therefore leads to higher holding torques of the joint, behavior analogous to that observed in normal BSJ. In our design, though, the jamming tension also has the effect of causing a change in the geometry of the joint, as the compliant elements shorten in length, while their cross-sectional area increases. This change alters the bending stiffness of the beam formed by the compliant elements and results in an increase in the perceived rotational stiffness of the joint.

\subsection{Manufacturing}

The rigid parts of the finger were printed using PolyTerra (\textit{Polymaker}) Polylactic Acid (PLA) filament, whilst Filaflex (\textit{Recreus}) TPU of shore hardness 83A was used for the compliant components. The latter material was selected as a good compromise between compliance and ease of printing, when compared to TPU 60A and 95A (Filaflex by \textit{Recreus}). The softer material proved difficult to print, leading to poor surface finish and geometrical accuracy of the final objects. In addition, its low Young's Modulus resulted in reduced stiffness range, which saturated even at low jamming tensions. In contrast, TPU 95A provided ease of printing, but required significant jamming forces to achieve any noticeable change in stiffness.

All parts were printed on an Ender 3 Plus S1 3D (\textit{Creality}). The bead tops and bottoms were FDM printed in TPU with 100\%\ infill to prevent the infill pattern from influencing the beads' bending behavior during deformation. In addition, an upright orientation was chosen when printing, to exploit the increased friction provided by the layer ridges, thereby increasing holding torque for a given jamming tension \cite{bead}. The PLA parts were printed using the default settings for CURA 5.6.0 (\textit{Ultimaker}). 

\section{Results and Discussion}
\label{sec:res}
\subsection{Experimental Testing Platform}
\begin{figure}[t]
    \centering
    \includegraphics[width=\columnwidth]{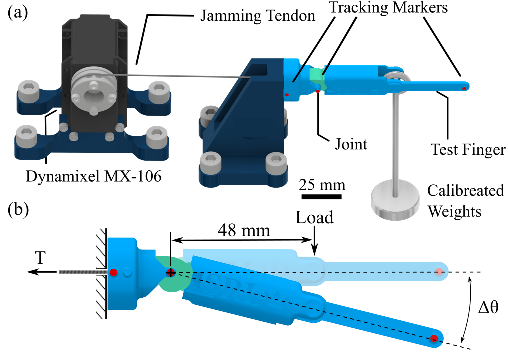}
    \caption{Overview of the experimental setup. (a) Each test sample consists of a single VS joint with markers for visual tracking. Jamming tension is applied through a steel tendon actuated by a Dynamixel MX-106 controlled in torque, while calibrated weights are used to provide external torque on the joint being tested. (b) Free body diagram of the test sample highlighting the rotational displacement $\Delta \theta$ used in the calculations throughout the study.}
    \label{fig:exp}
    \vspace{-1em}
\end{figure}

When characterizing the proposed CJJ mechanism, a single interphalangeal joint was isolated and tested, as seen in Fig. \ref{fig:exp}a. The jamming tension was provided through a pulley by a Dynamixel MX-106 (\textit{Robotis}) with torque control enabled. The phalanx and joint being tested were aligned with the top of the pulley, so that, in resting conditions, the jamming tendon run horizontally from the pulley, through the assembly, to the tip of the phalanx. Calibrated weights were applied at a distance of 48\,mm from the center of the joint (Fig. \ref{fig:exp}b). Red markers were placed at key locations along the test sample, to enable camera tracking during the experiments. Furthermore, as visible in the figure, a protrusion was added to the design on the phalanx to increase inter-marker spacing and improve tracking accuracy and angle measurements.

\subsection{Holding Torque Experiment}

The conventional method for benchmarking BSJ designs consists in evaluating their holding torque, which measures what forces the beaded system can withstand before friction between the joints is overcome, leading to relative motion between the beads and resulting plastic deformation of the structure \cite{bead, OGBead}. However, due to the compliance of our VS mechanism, this method could not be fully replicated. Instead, failure for our CJJ joints was defined as either a visible slip between the beads or the attainment of the full range of motion at constant load. 

In our experiments, torque values of 0.01\,Nm, 0.25\,Nm. 0.5\,Nm, 0.75\,Nm, 1\,Nm were output by the Dynamixel motor to generate jamming tensions of 1\,N, 17.86\,N, 35.71\,N, 53.57\,N, 71.4\,N. For every tension, calibrated weights of 10\,g were incrementally added to the tip of the finger until failure was observed, determining the resulting holding torque value. This procedure was repeated three times for each tension level, with a 10\,s interval between steps to allow sufficient time for the system to settle under the applied weight and to observe any potential joint failure.

\begin{figure}[h]
    \centering
    \includegraphics[width=\columnwidth]{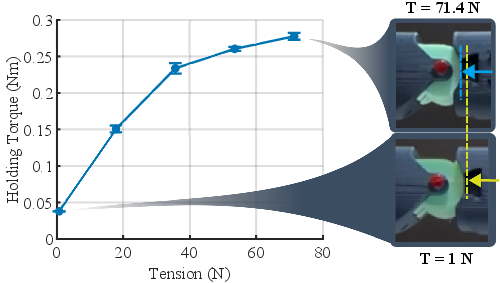}
    \caption{Results of the holding torque experiments displaying how an increase in tension affects the amount of torque the joint is able to withstand before failure. The non-linear relationship visible in the chart is due to the changing bead shape at increasing jamming tension. Errorbars in the plot represent the standard deviation obtained over three repetitions.}
    \label{fig:HT}
\end{figure}

Previous research investigating rigid beads has demonstrated how holding torque is directly influenced by the radius of the bead itself \cite{OGBead}. As previously stated, though, in our case a change in jamming tension also has the effect of altering the bead radius through elastic deformation of the compliant regions of the bead. Our results, shown in Fig. \ref{fig:HT}, display a non-linear relationship between tension and holding torque, unlike in rigid beads \cite{OGBead}: an initial steep change is followed by a plateauing region. This suggests that the change in radius due to deformation has an influence on the holding torque of the CJJ joint.

\subsection{CJJ Mechanism Stiffness and Hysteresis}
For this experiment, jamming tensions of 1$\,$N, 20$\,$N, 40$\,$N, 60$\,$N and 80$\,$N were used, with the maximum value dictated by the safe upper force range for the system. Load was applied to the tip of the phalanx using calibrated weights from 0$\,$g to 80$\,$g to ensure no failure for the 1$\,$N tension. During each test, the jamming tension was kept constant, while the external load was increased and then decreased in 20\,g steps, to assess for hysteretic behaviors of the joint.  For each loading condition, a pause of 2$\,$s was performed to let the system fully stabilize. Each test was repeated three times for three identical joint samples. To assess the stiffness behavior of the joints, the rotational displacement from a given load applied was used to calculate the rotational stiffness as the ratio between the torque provided by the tip load and the rotational displacement of the phalanx $\Delta \theta$.

\begin{figure}[h!]
    \centering
    \includegraphics[width=\columnwidth]{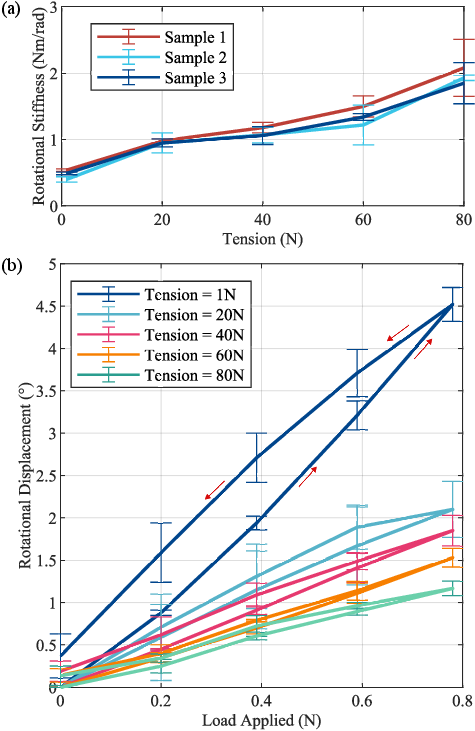}
    \caption{Joint stiffness characterization results. (a) Rotational Stiffness of the joint measured against the increase in tension applied to the jamming tendon. Errorbars in the plot represent the standard deviation obtained over three repetitions. (b) Chart showing the hysteresis behavior for Sample 1.The loops follow a counterclockwise direction, while errorbars represent the standard deviation obtained over three repetitions.}
    \vspace{-0.5 em}
    \label{fig:kvt}
  
\end{figure}

The average standard deviation across the three samples tested was 0.067\,Nm/rad, indicating good overall repeatability. As seen in Fig. \ref{fig:kvt}a, an increase in tension leads to an increase in the rotational stiffness, which varied from 0.48\,Nm/rad to 1.95\,Nm/rad. To determine whether the change in stiffness given an increase in jamming tension was statistically significant, the Wilcoxon Signed-Rank Test, suited for small sample sizes, was employed and each stiffness value at a given step was compared to the values in the next tension increase step. The test resulted in the null hypothesis being rejected with a 5\%\ significance level, meaning the tension-induced changes in stiffness were statistically significant.

Figure \ref{fig:kvt}b shows the results of the test without averaging results along the loading cycle. Significant hysteresis and higher standard deviations are observed for lower jamming tensions, with the system not returning to its initial condition with a tension of 1$\,$N. This is due to slippage in the joint during the test.  The situation drastically improves at higher jamming tension values, where manufacturing tolerances become the major source of uncertainty. The average residual deformation for 1$\,$N tension was 0.37$^{\circ}$, 0.12$^{\circ}$ for 20$\,$N, 0.19$^{\circ}$  for 40$\,$N, 0.09$^{\circ}$  for 60$\,$N and 0.07$^{\circ}$ for 80$\,$N.

Conducting a fatigue test for the joint served multiple purposes. First, it was essential to determine whether this mechanism could perform consistently over extended periods of use. Second, the test aimed to assess whether the materials used in the joint degrade over time.

\begin{figure}[ht]
    \centering
    \includegraphics[width=\columnwidth]{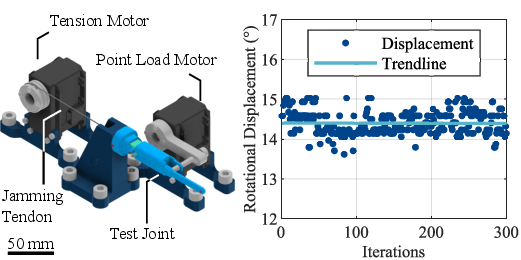}
    \caption{Experimental setup for the fatigue test. A second servo motor is introduced to directly apply torque to the joint by simulating a perpendicular point load on at the tip of test sample. The plot presents the final rotational displacement for a constant jamming tension of 89.29$\,$N and a point load of 16.4$\,$N, measured across successive iterations. The trendline equation is $y = -3.04\times10^{-5}x + 14.4$, confirming the absence of any significant change in stiffness of the joint.}
    \label{fig:fatigue}
\end{figure}

As displayed in Fig. \ref{fig:fatigue}, the experimental setup used to characterize the joint was modified, with a second Dynamixel MX-106, controlled in torque and coupled to the test sample, providing the point load at the tip of the phalanx. The axis of this motor was aligned with that of the joint being tested and its positional data was used to determine the angular displacement of the joint. A jamming tension of 89.29$\,$N and a point load of 16.4$\,$N were used during the trials. A large jamming tension was selected to maximize the stress on the system, while its value was capped to avoid the jamming tendon snapping mid-trial (behavior observed for tensions higher than 90\,N). The point load value was set to the maximum possible without incurring in slippage within the joint tested. The procedure for the experiment was as follows: (1) The point load servo positioned the finger at a perfectly straight configuration (0 degrees rotation around the joint). (2) Tension was applied to the jamming tendon. (3) Constant force was applied at the tip of the finger through the lever connected to the second servo motor. (4) Once equilibrium was achieved, the servo's recorded rotation was logged. (5) The point load was removed. (6) The tension was released. (7) The point load servo returned the joint to its straight configuration to prepare for the next trial. A total of 300 identical cycles were repeated and the final displacement of the joint in each trial was recorded. 

As visible in the fig. \ref{fig:fatigue}, a trendline fitted to the experimental data points has a negligible slope value of $-3.04\times10^{-5}$, meaning no significant changes in the behavior of the joint were detected. The calculated average standard deviation over the trials was 0.29 degrees. The variability observed in the results can be explained through the positional resolution of $\pm$0.33 degrees of the Dynamixel MX-106, as well as the resolution on the controlled output torque.

\subsection{Experimental validation in a peg-in-hole task}

\begin{figure*}[h]
    \centering
    \includegraphics[width=\textwidth]{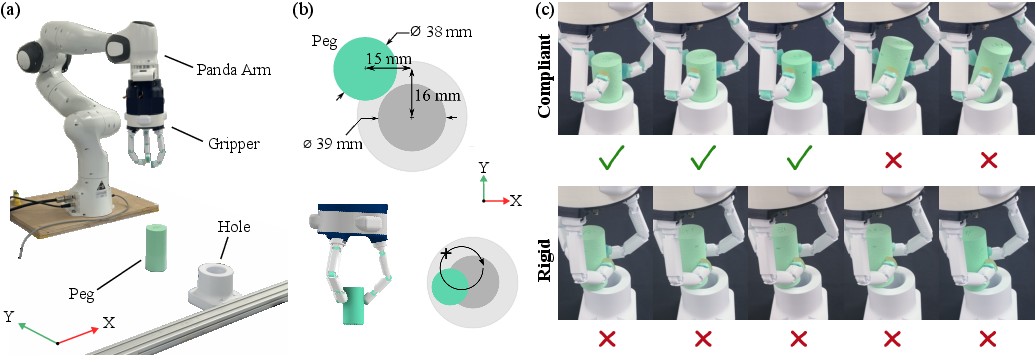}
    \caption{Peg-In-Hole validation experiment. (a) Experimental setup with Panda Arm. (b) The peg approaches the hole with a pre-specified misalignment. A circling maneuver is then used to try and line up the peg with the hole. The elements of the figure are not to scale (c) Experimental results comparing the rigid and compliant versions of the gripper, with `\ding{55}' representing failed peg insertion and `\checkmark' representing success. As can be seen for the compliant gripper, 60\%\ of attempts were successful.}
    \label{fig:val}
\end{figure*}

To demonstrate the advantages of passive residual compliance in adapting to uncertainties, a peg-in-hole task with intentional misalignment between the peg and the target hole was carried out. Two VS gripper systems were directly compared: a compliant three-fingered gripper utilizing CJJ to retain passive compliance and a three-fingered gripper based on rigid BSJ. The rigid gripper was identical to the compliant gripper, except that the TPU components were replaced with rigid PLA parts.  The peg used had a 1\,mm overall clearance with the hole, making the task challenging, to emphasize differences in the ability of rigid and compliant grippers to compensate for uncertainties during task execution. 

As shown in Fig. \ref{fig:val}a, both grippers were mounted on a Panda Arm (\textit{Franka Emika GmbH}) for maneuvering. SG90 servos (TowerPro) actuated the fingers with tendons, while a Dynamixel MX-106 enabled jamming on all three fingers simultaneously through a pulley (25\,mm radius) housing a separate helical grove for each tendon. The procedure for the experiment was as follows (as seen in supplementary video): (1) The peg (33.1$\,$g) was placed in the gripper. (2) A constant torque of 2.0\,N/m, split equally among the three fingers, was applied to stiffen them. (3) The gripper approached the hole at the pre-specified misalignment position. (4) The gripper performed a circular motion with a radius of 50\,mm and a speed of 0.1$\pi \, \mathrm{rad/s}$ (illustrated in Fig. \ref{fig:val}b). (5) After 40 seconds of circular motion, corresponding to two rotations, the gripper moved directly downward to attempt peg insertion. Success rates and failure modes were recorded across trials, capturing key performance differences between compliant and rigid grippers. The rigid gripper was not able to place the peg in the hole in any of the five trials, while the CJJ one had success in three out of the five tests. The observed 60\% increase in task success rate can be ascribed to the impact absorption capability afforded by the passive residual compliance within the joint (Fig. \ref{fig:val}c). Additionally, the compliant fingers maintain a downward force on the peg, helping it align with the hole, whereas the rigid fingers cannot adapt to misalignment and impact force, instead slipping off the peg during the impact.

\section{Conclusion}
\label{sec:end}
In this study, we introduced \textit{Compliant Joint Jamming} as a method for achieving passive residual compliance in an anthropomorphic finger design to improve dexterity. The proposed design, alternating rigid and compliant elements, can replicate the behavior of human fingers, with motion and stiffness variation localized at the interphalangeal joints \cite{postu}. By integrating compliance in the traditional BSJ framework, we achieved a stiffness range spanning from 0.48$\,$Nm/rad at 1$\,$N tension to 1.95$\,$Nm/rad at 80$\,$N tension. This results in a 4-fold increase in stiffness. A key innovation in the proposed design is the incorporation of a notch in the bead structure, which significantly enhances the range of motion of the fingers from 45$^{\circ}$ to 72$^{\circ}$ allowing for a more accurate recreation of the range of motion of human fingers.

A limitation of this mechanism is the hysteresis observed at low tensions due to bead slipping. While using materials with a higher coefficient of friction for the compliant beads could mitigate this issue, such materials are often challenging to 3D print with. An alternative would be surface texturing either during printing or at the post-processing stage. In future, developing a model to explain changes in holding torque and stiffness based on changes in geometry of the bead as a function of the jamming tension will provide a theoretical framework to better predict and control stiffness characteristics, supporting precise design optimization and enhancing performance predictability for various applications. Even at the current stage, our validation through the proposed peg-in-hole task demonstrates the importance of including passive residual compliance in VS designs to compensate for uncertainties in task execution while minimizing sensor and control needs. In particular, the ability of the gripper to absorb impacts without losing its grasp on the peg proved to be vital in increasing task success rate.

 
%

\section*{ACKNOWLEDGMENT}

The authors would like to thank Giammarco Caroleo for his assistance with the validation experiment.

\bibliographystyle{unsrt}
\bibliography{references}

\end{document}